\documentclass[10pt,twocolumn,letterpaper]{article}

\usepackage[pagenumbers]{cvpr} % To force page numbers, e.g. for an arXiv version

\usepackage{graphicx}
\usepackage{amsmath}
\usepackage{amssymb}
\usepackage{booktabs}
\usepackage{siunitx}

\usepackage[pagebackref,breaklinks,colorlinks]{hyperref}

\usepackage[capitalize]{cleveref}
\crefname{section}{Sec.}{Secs.}
\Crefname{section}{Section}{Sections}
\Crefname{table}{Table}{Tables}
\crefname{table}{Tab.}{Tabs.}

\def\cvprPaperID{2} % *** Enter the CVPR Paper ID here
\def\confName{OmniCV}
\def\confYear{2023}

\begin{document}

%%%%%%%%% TITLE - PLEASE UPDATE
\title{Applications of Deep Learning for Top-View Omnidirectional Imaging: A Survey}

\author{Jingrui Yu \qquad Ana Cecilia Perez Grassi \qquad Gangolf Hirtz \\
Chemnitz University of Technology, Germany\\
\tt\small \{jingrui.yu, ana-cecilia.perez-grassi, g.hirtz\}@etit.tu-chemnitz.de
}
\maketitle

%%%%%%%%% ABSTRACT
\begin{abstract}
  A large field-of-view fisheye camera allows for capturing a large area with minimal numbers of cameras when they are mounted on a high position facing downwards. This top-view omnidirectional setup greatly reduces the work and cost for deployment compared to traditional solutions with multiple perspective cameras. In recent years, deep learning has been widely employed for vision related tasks, including for such omnidirectional settings. In this survey, we look at the application of deep learning in combination with omnidirectional top-view cameras, including the available datasets, human and object detection, human pose estimation, activity recognition and other miscellaneous applications.
\end{abstract}

%%%%%%%%% BODY TEXT
\section{Introduction}
\label{sec:intro}

Omnidirectional cameras have the advantage of being able to capture a wide field of view (FOV). However, their projection models introduce a large distortion into their images. For this reason, computer vision methods developed for perspective images are not suitable for omnidirectional ones.
In the last decade, computer vision has experienced a great advance thanks to the development of deep neural networks and the availability of large databases. However, this advance has focused almost exclusively on perspective images, both in the development of architectures and in the collection and annotation of data. It has not been until recent years that deep learning has begun to reach omnidirectional image processing, by collecting datasets and adapting existing architectures or developing new ones for this type of image.

Omnidirectionality can be achieved by using catadioptric, dioptric or polydioptric cameras. Catadioptric cameras combine a normal camera with a shaped mirror \cite{nayar1997catadioptric,gaspar2000vision,winters2000omnirobot}. This mirror provides omnidirectionality as a surround-view, but the camera itself occludes the central part of the image. This problem is solved by dioptric cameras, which use a fisheye lens instead of a mirror. Finally, polydioptric cameras capture a spherical field of view by combining multiple cameras in a setup \cite{8770080,li2019stitching}.

Especially dioptric cameras are gaining attention in many applications because they are simple and inexpensive. Depending on the task, these cameras can be mounted with a frontal view, as for example in driving applications \cite{Rashed_2021_WACV,yahiaoui2019fisheyemodnet,yogamani2019woodscape,Kumar_2021_WACV}, with a vertical view as in teleconference applications \cite{otsuka2009realtime} or with a top view as in surveillance applications \cite{li2019supervised,nguyen2016real,minh2021arpd,laurendin2021hazardous}. Also, their use for 3D-reconstruction, using one or more cameras, increases in the recent years \cite{6942637,Li2019roomreconstruction,won2020omnislam}.
In this survey we focus on deep learning algorithms developed for fisheye images captured from a top view. This kind of images are essential in surveillance and Ambient Assisted Living (AAL) applications \cite{seidel2018auxilia}, where the main research areas include person and object detection and human pose estimation.

Although there exist other surveys that focus on omnidirectional fisheye images, such as \cite{kumar2023fisheyesurvey,silveira2022survey,ai2022omnisurvey}, they mostly discuss the frontal view.
Other surveys of top-view imaging \cite{liciotti2017survey,ahmad2019survey} concentrate only on one application of the top-view perspective and do not specify the usage of omnidirectional cameras.
The methods surveyed are mostly confined to classical computer vision algorithms.
Therefore, we consider our survey essential for grasping the trend in applications of the combination of top-view fisheye imaging and deep learning.

This paper is organized as follows:
in \cref{sec:setup} we describe the camera geometry and the top view setup in detail.
The available omnidirectional datasets are presented in \cref{sec:datasets}.
Sections \ref{sec:detection}, \ref{sec:pose} and \ref{sec:other} cover object detection, pose estimation, human activity recognition and other miscellaneous applications.
We conclude the survey in \cref{sec:concl}.

\section{The omnidirectional top-view setup}
\label{sec:setup}
\begin{figure}[tb]
    \centering
    \subcaptionbox{The top-view omnidirectional set-up in a one-room apartment. The red dot in the center illustrates the position of the camera.}[\linewidth]{\includegraphics[width=\linewidth]{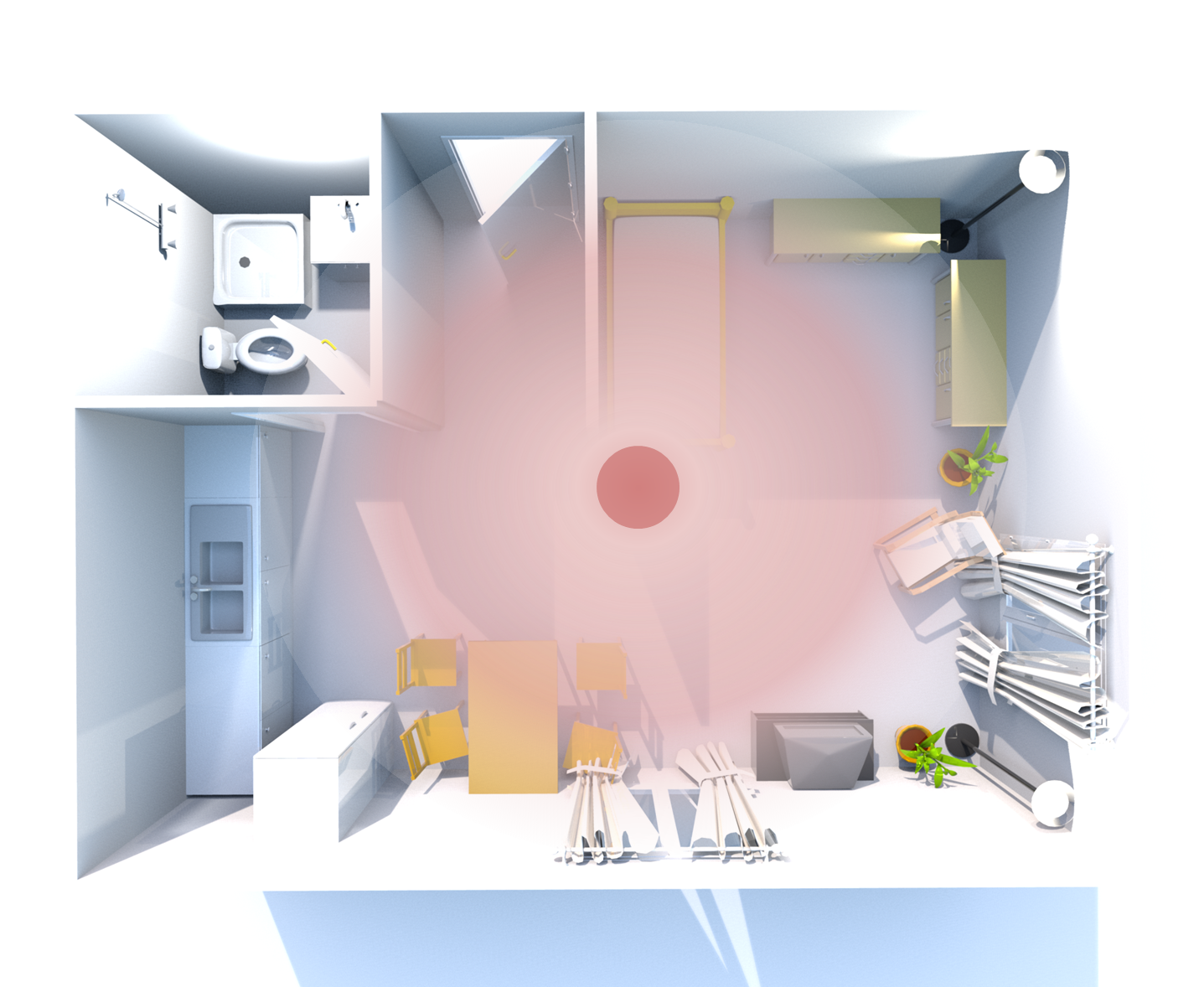}}
    \par\medskip
    \subcaptionbox{Example output of the set-up in a synthetic environment.}[\linewidth]{\includegraphics[width=0.85\linewidth]{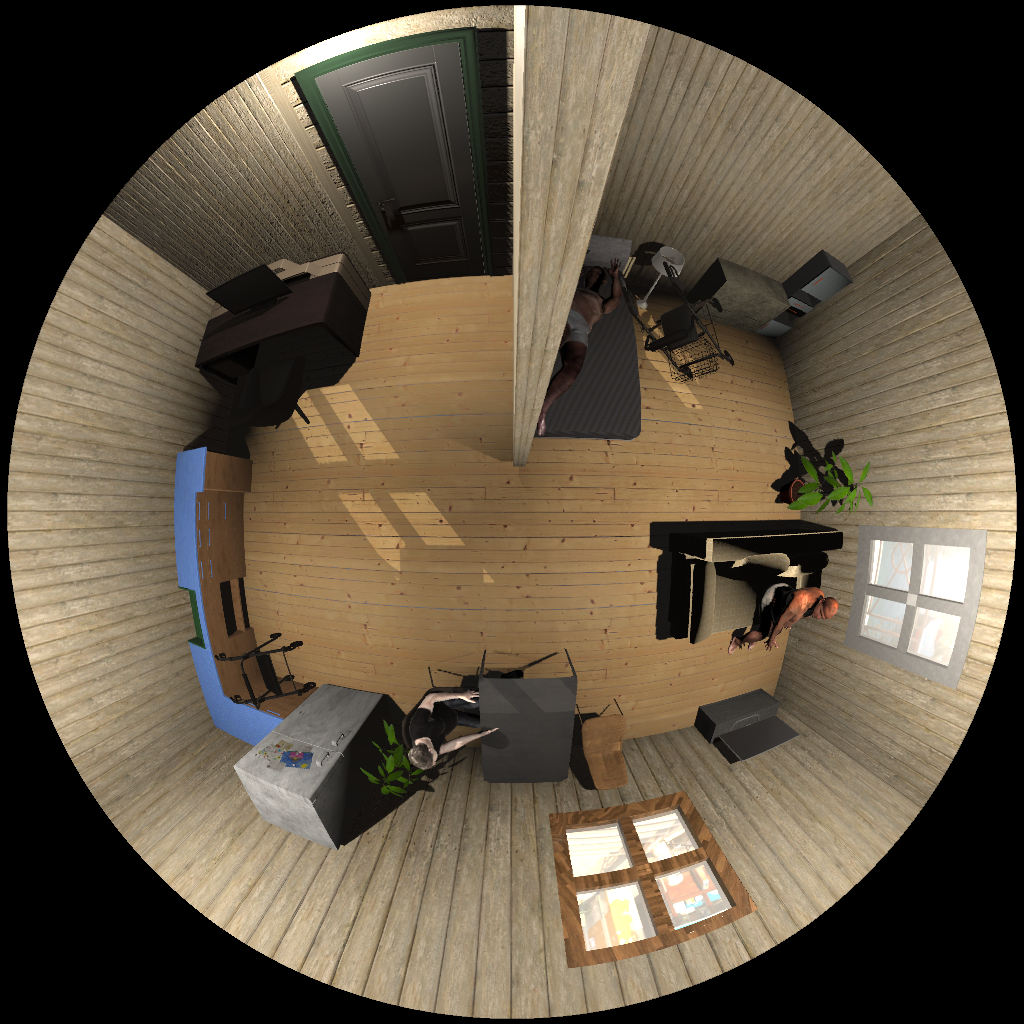}}
    \par\medskip
    \caption{The top-view omnidirectional set-up and its output.}
    \label{fig:setup}
\end{figure}

As introduced in \cref{sec:intro}, this survey focuses of the top-view omnidirectional vision utilizing one or multiple dioptric cameras.
The camera or the camera rig consisting of multiple cameras is usually hung on the ceiling near the center of the room.
Figure \ref{fig:setup} shows an example of such setting.
Do not confuse this with the top view or bird's-eye-view in autonomous driving applications~\cite{synwoodscape}, which is synthesized from the surround view images.

\begin{figure}
    \centering
    \includegraphics[width=0.9\linewidth]{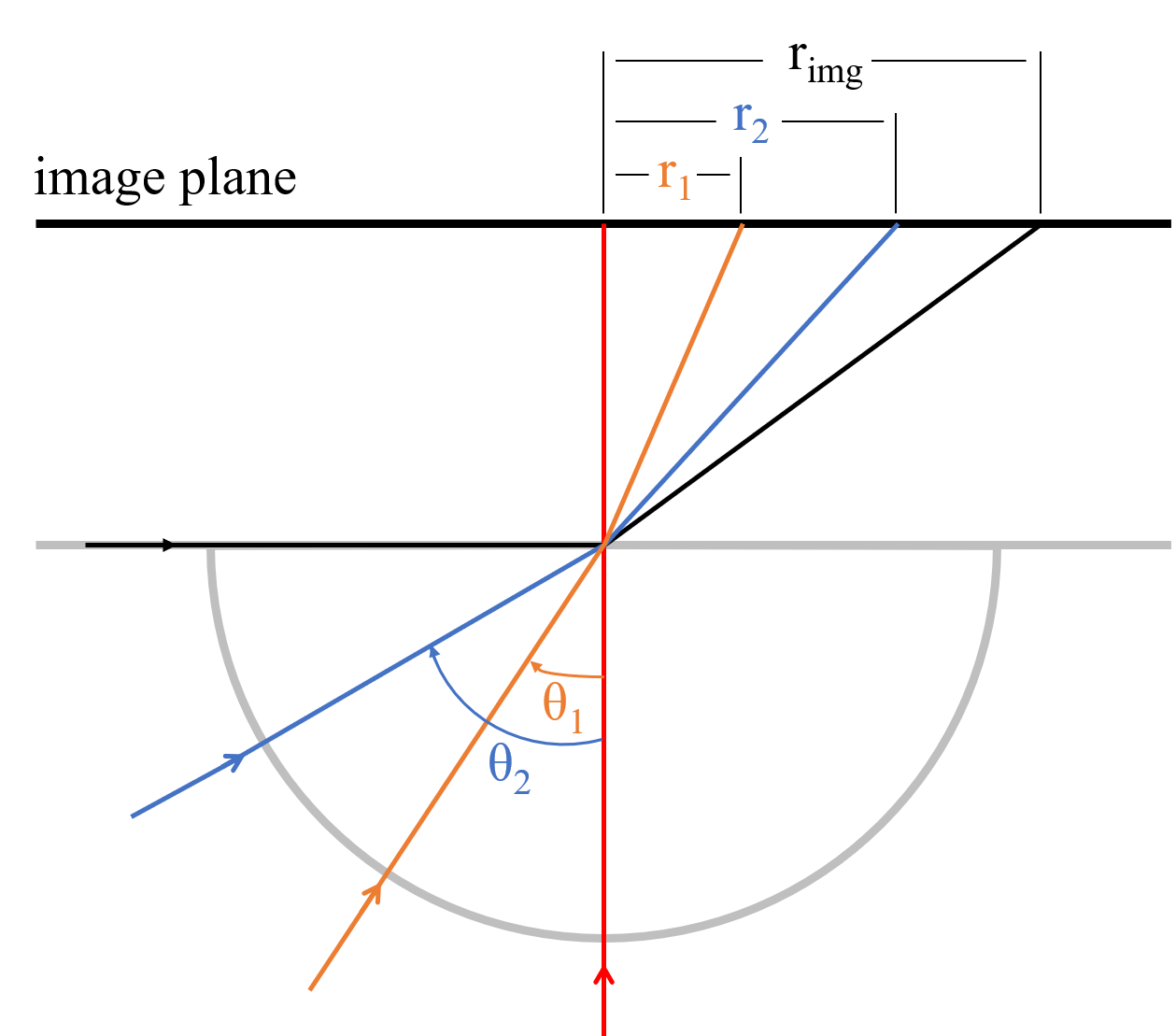}
    \caption{The equidistant projection. The red ray is the principle axis. The black ray is from the maximum visible angle and \(r_{img}\) is the radius of the image area.}
    \label{fig:equiproj}
\end{figure}

To utilize the advantages and tackle the shortcomings of this set-up, we need to understand the model of fisheye camera.
An ideal fisheye lens can be described with the equidistant projection in \cref{eq:equiproj}.
\(\theta\) is the angle between the principal axis and the incoming ray, \(r\) is the distance between the image point and the principal point, and \(f\) is the focal length (see \cref{fig:equiproj}).
\begin{equation}
    r = f\theta
    \label{eq:equiproj}
\end{equation}
There are other projection models that are less frequently used, see \cref{tab:fisheyeproj}.

\begin{table}
    \centering
    \caption{Other projection models used for fisheye cameras besides equidistant projection.}
    \label{tab:fisheyeproj}
    \begin{tabular}{l|l}
        Projection type & Math. expression \\
        \midrule
        Equisolid & \(r=2f\sin(\theta/2)\) \\
        Stereographic & \(r=2f\tan(\theta/2)\) \\
        Orthographic & \(r=f\sin\theta\) \\
    \end{tabular}
\end{table}

A realistic fisheye camera is not perfect and require a more complex model for calibration and precise image unwrapping and odometry.
The publication \cite{kannala2006fisheyemodel} describes a generic model that enables the calibration of a fisheye camera with a single planar calibration pattern.
DeepCalib presents the possibility of using DL for acquiring calibration parameters of fisheye cameras~\cite{bogdan2018deepcalib}.

An omnidirectional camera can be built with a normal CCD or CMOS camera and a fisheye lens~\cite{rehman2022hemispherical}.
Besides, there are commercial products from various companies, such as the Quasar\texttrademark\ Hemispheric Mini-Dome by TELEDYNE FLIR, the HemiStereo\texttrademark\ DK1 and NX by 3DVisionlabs, the panoramic series by HIKVISION, the IP Fisheye series by ABUS, the C71 and Q71 by MOBOTIX, the FE series by VIVOTEK, etc.
They can be most easily found under the term ``hemispherical camera'' with a search engine.

\section{Datasets}
\label{sec:datasets}

In deep learning approaches, the availability of quality data and ground truth annotations is essential for the training process.
Today, models of almost every architecture can be obtained with weights trained on popular large-scale image datasets such as MS COCO~\cite{lin2014coco}, ImageNet~\cite{ILSVRC15,yang2019fairer}, etc.
This enables users to avoid long and costly trainings from scratch and to facilitate the transfer learning ability of neural networks to adapt these existing models to new domains and new tasks. 
However, the images in these datasets are mostly collected with a perspective camera from a frontal view.
Therefore, the pictured objects present a different appearance from those in top-view omnidirectional images, especially when they are close to the camera.
This prevents not only the direct use of these models in such images, but also makes transfer learning extremely difficult, if a large amount of omnidirectional data for fine-tuning is not available.
In the early 2010s, omnidirectional image datasets were scarce and insufficient for training or fine-tuning complex architectures.
However, with the increasing interest in using omnidirectional cameras, the first real-world and synthetic datasets of top view fisheye images were created.
Unlike general-purpose image datasets for classification, object recognition or segmentation, which are collected on the internet and have great intra-dataset variability, omnidirectional datasets are mostly recorded in a specific setting for a specific task.
Therefore, they are usually continuous sequences extracted from videos, and the variability between images is lower.
In the following subsections, we present and describe these datasets.
\cref{tab:datasets} summarizes their technical characteristics.
Links to the datasets in this chapter are accessed on March 17th, 2023.

\begin{table*}[tb]
    \centering
    \caption{Technical attributes of omnidirectional image and video datasets. Types include real-world (R), synthetic (S) or hybrid (R+S). Res. stands for resolution and the unit is megapixels (MP). Year indicates the year of publication, not the year of appearance.}
    \label{tab:datasets}
    \small
    \begin{tabular}{lcrrllc}
        \toprule
        Dataset & Type & \# of frames & Res. (MP) & Annotations & Classes & Year \\
        \midrule
        Bomni & R & 10,340 & 0.3  & bbox, tracking ID, actions & 9 actions & 2012 \\
        HDA (\textit{Cam 02}) & R & 1,388 & 0.3 & bbox & person & 2013 \\
        LMS & R+S & 515 & 1.2 & -- & -- & 2016 \\
        PIROPO & R & 111,283 & 0.48 & head point, bbox (3rd party) & person & 2021 \\
        MW-18Mar & R+S & 14,040 & 1.1 to 2.2 & bbox, rotated bbox (3rd party) & person & 2018 \\
        HABBOF & R & 5,837 & 4.2 & rotated bbox & person & 2019\\
        CEPDOF & R & 25,504 & 1.1 to 4.2 & rotated bbox & person & 2020 \\
        WEBDTOF & R & 10,544 & 0.6 to 5 & rotated bbox & person & 2022 \\
        FRIDA & R & 18,318 & 4.2 & rotated bbox, person ID & person & 2022 \\
        DEPOF & R & 3,594 & 4.2 & bbox, point location & person & 2023 \\
        FES & R & 301 & 2.8 & bbox, instance mask & 6 classes & 2020 \\
        360Action & R & 784 clips & 8.3 & actions per video clip & 19 actions & 2020 \\
        FRailTRI20\_DOD & R & 44,099 & 0.93 & temporal anomalies & 7 anomalies & 2020 \\
        OSD & R+S & 39,200 & 1.0 & bbox & person & 2021 \\
        THEODORE & S & 100,000 & 1.0 & bbox, segmentation \& instance mask & 14 classes & 2020 \\
        THEOStereo & S & 31,250 pairs & 1.0 & depth map & -- & 2021 \\
%        OmniFlow & S & 23,653 & 4.2 & optical flow & person & 2021 \\
        % \midrule
        % EgoCap & R & 75,214 & 0.3 & 2D keypoints & 17 keypoints & 2016 \\
        % Mo\textsuperscript{2}Cap\textsuperscript{2} & S/H & 530k & & 2D \& 3D keypoints & 15 keypoints & 2019 \\
        % xR-EgoPose & S & 380k & 1.0 & 2D \& 3D keypoints, seg. mask, actions & & 2019 \\
        % EgoPW & R & 318k & 1.0(ego), 2.0(ex) & 3D keypoints & 15 keypoints & 2022 \\
        \bottomrule       
    \end{tabular}
\end{table*}

\subsection{Real-world datasets}
\label{subsec:realD}

The \textbf{Bomni Database}\footnote{\url{https://www.cmpe.boun.edu.tr/pilab/pilabfiles/databases/bomni/}} (Boğaziçi University Multi-Omnidirectional Video Tracking Database)~\cite{demiroz2012} is one of the earliest datasets of omnidirectional fisheye camera images.
Although the authors list a few other datasets, they are not available anymore at the time of this review.
Bomni DB is recorded for the purpose of human tracking in indoor scenes.
Two fisheye cameras, one mounted on the ceiling and the other on a side wall, are used to simultaneously record two scenarios with a
resolution of \(640\times 480\) pixels and a frame rate of \SI{8}{fps}.
Scenario 1 shows a single subject entering a room and performing six different actions before leaving.
For this scenario a total of 10 videos with 5 different subjects are recorded.
Scenario 2 presents 36 videos of multiple persons interacting in the same room.
For this scenario a total of five actions are defined.
The dataset provides tracking IDs, bounding boxes for moving subjects and action labels as annotations, which are given in vatic~\cite{vondrick2013} format.
It is to be noted that a portion of the annotations are generated from automatic tracking and interpolation. 
This often results in slight misalignment between the subject and its bounding box.
Additionally, Bomni DB lacks labels for quasi-static persons in the scene.

\textbf{HDA Person Dataset}\footnote{\url{https://vislab.isr.tecnico.ulisboa.pt/hda-dataset/}}~\cite{figueira2014} is a dataset for surveillance.
Most of the image data are recorded by classic surveillance cameras, but the sequence \emph{Cam 02} is recorded by a fisheye camera mounted on the ceiling of an elevator waiting area.
The sequence has a resolution of \(640\times 480\) and a frame rate of \SI{5}{fps}.
In total 9819 frames are recorded.
Bounding boxes of persons are provided in this dataset.
Heavy motion blur is present throughout this recording.

\textbf{PIROPO database}\footnote{\url{https://sites.google.com/site/piropodatabase/}} (People in Indoor ROoms with Perspective and Omnidirectional cameras)~\cite{del2021} is recorded simultaneously using a ceiling-mounted fisheye camera and a normal perspective camera.
The scenes consist of a single person or multiple people walking, standing or sitting in a room.
There are no interactions between the persons.
It is a large scale dataset with over 100,000 annotated frames and a number of unannotated frames.
The annotation is provided in the form of points, which mark the head positions of the persons in the image.
In the work \cite{yu2019omnipd}, the authors mention they down-sampled the original dataset with annotations and manually annotated the resulting dataset with bounding boxes for the persons.

% 32892 (Omni1A) + 15145 (Omni1B) + 39820 (Omni2A) + 23426 (Omni3A)

\textbf{MW-18Mar Dataset}\footnote{\url{https://www2.icat.vt.edu/mirrorworlds/challenge/index.html}} from Mirror Worlds Challenge is an indoor top-view fisheye video dataset which consists of 30 videos an 13k frames. The original dataset are annotated with axis-aligned bounding boxes. 
For tracking purpose there are also annotated track trajectories.
There are 3 main scenarios in this dataset: an observation room, a hallway and a synthetic scene of an observation room.
The train set of this dataset is later annotated with rotated bounding boxes by the authors of \cite{duan2020rapid} and is named \textbf{MW-R}\footnote{\url{https://vip.bu.edu/projects/vsns/cossy/datasets/mw-r/}}.

\textbf{Tamura \etal} annotated Bomni, PIROPO and MW-18Mar datasets with rotated bounding boxes for their work \cite{tamura2019}.
The annotation files are in Pascal VOC format~\cite{everingham2010pascal} and available online\footnote{\url{https://github.com/hitachi-rd-cv/omnidet-rotinv}}.

\textbf{HABBOF} (Human-Aligned Bounding Boxes from Overhead Fisheye Cameras)~\cite{li2019supervised}, \textbf{CEPDOF} (Challenging Events for Person Detection from Overhead Fisheye Images)~\cite{duan2020rapid} and \textbf{WEBDTOF} (In-the-Wild Events for People Detection and Tracking from Overhead Fisheye Cameras)~\cite{tezcan2022} are datasets collected by the Visual Information Processing Laboratory of Boston University\footnote{\url{https://vip.bu.edu/projects/vsns/cossy/datasets}}.
HABBOF provides two indoor scenes of 5837 frames.
The annotations are given as bounding boxes aligned to the human body, which appear mostly in line with the radial axis of the omnidirectional image.
CEPDOF is an extension of HABBOF.
It provides 8 video sequences of different levels of crowdedness under different lighting conditions.
Unlike any earlier datasets, which are recorded in controlled settings, WEBDTOF is recorded in real-life situations.
14 scenes are recorded with different cameras and lens to form 16 videos.
Thus, it covers common difficulties presented in real life: occlusions, camouflage, cropping, tiny people, non-circular FOV and children.
The same research group also presents two datasets for other applications:
\textbf{DEPOF} (Distance Estimation between People from Overhead Fisheye cameras)~\cite{lu2023DEPOF} and \textbf{FRIDA} (Fisheye Re-Identification Dataset with Annotations)~\cite{cocbas2022FRIDA}.
DEPOF provides 3,526 frames for calibration purpose and 68 frames for training and testing of a person distance measurement method.
FRIDA is destined for person re-identification, but rotated bounding boxes are also available.

\textbf{FES}\footnote{\url{https://www.tu-chemnitz.de/etit/dst/forschung/comp_vision/datasets/fes/}} (Fisheye Evaluation Suite)~\cite{scheck2020learning} is an indoor dataset.
It differs greatly from the afore mentioned datasets by providing bounding boxes and instance segmentation masks for 6 classes: person, TV, table, armchair, chair and wheeled walker.
The disadvantage of this dataset is its relative small size at only 301 frames.

\textbf{360action}\footnote{\url{https://github.com/ryukenzen/360action}}~\cite{li2020weakly} is a dataset for action recognition in the form of video clips. 
In each clip, which has the length of 6 to 10 seconds at \SI{30}{fps}, a number of subjects perform daily actions by themselves or with interactions.
Action labels are given for each video, without specifying the subject.

\textbf{FRailTRI20\_DOD} (French Rail Technological Research Institute Door Obstacle Detection 2020)~\cite{laurendin2021hazardous} was created specifically for the surveillance of door areas of trains. The images were captured at \SI{20}{fps}. Seven anomalies regarding train doors and passengers were performed by five actors. The authors defined a set of annotations regarding instance positions and displacement, door state, hazardous events, pedestrian actions and combine them into a temporal segmentation of an event. The dataset as well as a meticulous annotation guide can be acquired by contacting the first author.

\subsection{Synthetic datasets}
\label{subsec:synthD}

Apart from the subset in MW-18Mar dataset, multiple synthetic datasets of omnidirectional images have been created for a variety of purposes.

\textbf{LMS Fisheye Dataset}\footnote{\url{https://www.lms.tf.fau.eu/research/downloads/fisheye-data-set/}}~\cite{eichenseer2016} provides a variety of synthetic and real-world fisheye image video sequences, among which \textit{HallwayC}, \textit{LivingroomB}, \textit{Room}, \textit{HallwayB}, \textit{LivingroomA} and \textit{LivingroomC} are top view. 
These sequences do not have corresponding annotations.

\textbf{OSD}\footnote{\url{https://datasets.vicomtech.org/v4-osd/OSD_download.zip}} (Omnidirectional Synthetic Datasets)~\cite{aranjuelo2021key} is a dataset for person recognition and surveillance.
Therefore, it only provides annotations for persons in the form of segmentation masks and bounding boxes. 
The persons are small in size to simulate actual surveillance situations where a large area is monitored by one omnidirectional camera.
Besides Omnidirectional images it also provides rectified images. 

\textbf{THEODORE}\footnote{\url{https://www.tu-chemnitz.de/etit/dst/forschung/comp_vision/datasets/theodore/}} (synTHEtic tOp-view inDoOR scEnes dataset)~\cite{scheck2020learning} provides a large-scale diverse dataset with annotations for semantic segmentation, instance segmentation and bounding boxes for object detection.
\textbf{THEOStereo}\footnote{\url{https://www.tu-chemnitz.de/etit/dst/forschung/comp_vision/datasets/theostereo/}}~\cite{seuffert2021study} is derived from THEODORE and aims to aid depth estimation using top-view fisheye cameras. It provides image pairs from two virtual cameras and the corresponding depth maps as ground truth annotations. The baseline of the stereo cameras is \SI{0.3}{\meter}.

\subsection{Other datasets}
\label{subsec:otherD}
Other datasets such as \textbf{DPI-T}\footnote{\url{https://github.com/zhengkang86/ram_person_id}} (Depth-Based Person Identification from Top View)~\cite{haque2016recurrent}, \textbf{TVPR}\footnote{\url{https://vrai.dii.univpm.it/re-id-dataset}} (Top View Person Re-Identification) dataset~\cite{Liciotti2017reid} does not use fisheye cameras, but the overhead viewpoint is nonetheless useful for training networks applied for omnidirectional images. These two datasets contain a depth channel in addition to RGB channels.
\textbf{PanopTOP31K}\footnote{\url{https://github.com/mmlab-cv/PanopTOP}}~\cite{Garau2021panoptop} is a semisynthetic RGB dataset from the top-view for view-invariant 3D human pose estimation.
This is the first large-scale dataset that features top-view human keypoints.
However, its low resolution at \(256\times256\) pixels and heavy artifacts affect its usability negatively.

\section{Person and object detection}
\label{sec:detection}

\subsection{Person detection}
\label{subsec:personD}

The datasets in \cref{sec:datasets} clearly show that the main application of omnidirectional cameras is person detection and tracking.
Using deep learning for person detection in omnidirectional images, typically a CNN-based object detector, faces a few obstacles.
Firstly, standing people appear in line with the radial axis of the image rather than upwards in images from side-mounted cameras.
Secondly, the equidistant projection of fisheye cameras results in considerable deformation of objects.
These two problems restricts the utilization of pre-trained models and reduces the effectiveness of transfer learning.
Additionally, when the person stands directly under the camera, it has a unique appearance that is unseen in normal perspective images.
We review the researches to see how these problems are progressively solved.
An overview is provided in \cref{tab:personD}.
Note that due to the lack of a common large scale dataset, researchers have used different datasets and metrics for the evaluation, therefore it is not possible to compare the performances to each other directly.
Thus, the performances are not listed in the table.

\begin{table*}[ht]
    \centering
    \caption{Person detection in Omnidirectional images}
    \label{tab:personD}
    \small
    \begin{tabular}{rl}
        \toprule
        Architecture & Main algorithm \\
        \midrule
        Nguyen \etal~\cite{nguyen2016real} &  AGMM-background subtraction + tiny YOLO \\
        OmniDetector~\cite{seidel2018improved} & Unwapping + YOLOv2 + NMS \\
        Li \etal~\cite{li2019supervised} & Rotating window + background subtraction + YOLOv3 \\
        Tamura \etal~\cite{tamura2019} & Rotation-invariant training + YOLOv2 + BBR \\
        OmniPD~\cite{yu2019omnipd} & Hybrid training + rotation augmentation \\
        RAPiD~\cite{duan2020rapid} & YOLOv3-based network + orientation prediction head + angle aware loss \\
        ARPD~\cite{minh2021arpd} & CenterNet + orientation prediction head + rotation aware loss function \\
        Haggui \etal~\cite{haggui2021human} & RAPiD + color histograms for tracking \\
        Wang \etal~\cite{wang2019mask} & Dual Mask R-CNN + image region separation + scene specific training \\
        GSAC-DNN~\cite{fuertes2022people} & 2D grid of simple CNN-classifiers \\
        Callemein \etal~\cite{callemein2019anyone} & Low resolution image + temporal interlacing kernel in YOLOv2 \\
        OmniDRL~\cite{pais2019omnidrl} & Deep Q-Network + camera calibration \\
        Wiedemer \etal~\cite{Wiedemer2022fewshot} & Faster-RCNN + few-shot training \\
        \bottomrule
    \end{tabular}
\end{table*}

\textbf{Nguyen \etal}~\cite{nguyen2016real} combined Adaptive Gaussian Mixture Model (AGMM)-based background subtraction and a simple CNN inspired by Tiny Yolo~\cite{redmon2016yolo} to perform pedestrian detection.
The network takes the foreground mask and single-channel grayscale images as input.
Their evaluation shows an AP of 0.86 at their house dataset.

Seidel \etal proposed \textbf{OmniDetector}~\cite{seidel2018improved}, which uses the camera calibration parameters to unwrap one omnidirectional image into 94 highly overlapping perspective images and then apply the pre-trained YOLOv2~\cite{redmon2017yolo9000} to detect persons.
The bounding boxes are projected back into the omnidirectional image using a look-up table (LUT).
Non-maximum suppression (NMS) is applied to the detections to eliminate overlapping bounding boxes and generate the final detection.
They achieved an AP@0.5IoU of 0.646 on PIROPO when using soft-NMS with Gaussian smoothing.
This method enables the use of CNN-based detectors without the need for collecting new data and training.
However, its shortcomings are obvious.
It has a large overhead, partly because of the transformations and partly because of the large amount of inferences for one image.
It cannot detect persons directly under the camera, since the network has not seen such examples.
Furthermore, it requires that the camera calibration parameters are known, which isn't feasible at all times.
Curiously, \textbf{Chiang \etal}~\cite{chiang2021efficient} used the same approach in 2021, with the only improvement of reducing the number of ROIs to eight.

\textbf{Li \etal} proposed in \cite{li2019supervised} to use a rotating rectangular focus windows to extract a part of the image, which will be rotated to maintain the upright direction of the person.
The maximum number of focus windows is 24.
Then the detection is performed with YOLOv3~\cite{redmon2018yolov3} and consecutive NMS.
The authors used background subtraction to identify regions of interest (ROI) where people are present and discard the focus windows without human activity, thus reducing the computational cost.
This method is tested on \mbox{HABBOF} with the F-score of 0.88.
The similarities with OmniDetector are the usage of multiple overlapping windows and that the detector does not need to be re-trained or fine-tuned.
But this method reduces the computational cost and does not require the camera parameters to be known.
However, it still does not address the person-under-the-camera problem, and it is by design not able to detect stationary persons.

Yu \etal took another approach and tried to achieve person detection directly in omnidirectional images with \mbox{\textbf{OmniPD}}~\cite{yu2019omnipd}.
They presented a training paradigm, by which omnidirectional images are combined with a dataset of normal perspective, in this case the PASCAL VOC dataset~\cite{everingham2010pascal}, to finetune a CNN-based object detector.
Random horizontal and vertical flipping and random 90-degree rotation was used as data augmentation to compensate for rotation variance in omnidirectional images.
Their best result was achieved with SSD~\cite{Liu2016ssd} at AP@0.5IoU at 0.863, albeit on their own dataset.

\textbf{Tamura \etal} tried to achieve pedestrian detection in omnidirectional images by training YOLOv2 with randomly rotated perspective images from COCO~\cite{tamura2019}.
\mbox{DPI-T} dataset was also used in the training.
To overcome YOLOv2's problem of generating overlapping bounding boxes, they proposed bounding box regression (BBR) based on mean shift clustering of the center points of bounding boxes.
A simple position-based bounding box angle determination was added to the refinement process.
The authors manually annotated MW-18Mar, PIROPO, Bomni and CVRG for evaluation, as mentioned in \cref{subsec:realD}.

Duan \etal proposed \textbf{RAPiD}~\cite{duan2020rapid}, a new YOLO-inspired network architechture, which predicts the rotation angle besides the usual position and size of the bounding boxes.
To train this network they added rotation-angle loss to the loss function of YOLOv3.
There network is first pre-trained on COCO, then finetuned on two of the three datasets they annotated (MW-R, HABBOF, CEPDOF) and tested on the remaining one.
They reached AP@0.50IoU of 0.967, 0.981 and 0.858 for MW-R, HABBOF and CEPDOF, respectively.
Their team further improved the performance by extending RAPiD with temporal information~\cite{tezcan2022}.
Minh \etal used the same strategy to  extend CenterNet to predict human aligned bounding boxes, and named their architecture \textbf{ARPD}~\cite{minh2021arpd}.
Their experiments show that their method reaches similar AP as RAPiD while doubling the inference speed.
\textbf{Haggui \etal} used RAPiD for initial detection and added tracking by using the color histograms~\cite{haggui2021human}.

The most recent development is \cite{Wiedemer2022fewshot}.
\textbf{Wiedemar \etal} proposed a few-shot adversarial training scheme for Faster-RCNN~\cite{ren2015fasterrcnn} so that a pre-trained detection model can be adapted for person detection in top-view omnidirectional images with less than 100 annotated training samples.
The techniques they used include loss coupling, global and instance level feature alignment.
Their method can achieve higher accuracy when the number of annotated samples are smaller than 100.
A key difference to previously mentioned methods is that this method is aimed at adapting an existing model with minimum amount of effort to a certain use case, instead of trying to create a model with maximum generalization power.
Therefore, cross-dataset evaluation by the author shows that the model loses generalization power when the number of training examples exceed 50.

Besides the common methods for detection, researches have experimented with other ways of person detection with special features.
\textbf{Wang \etal}~\cite{wang2019mask} proposed to use Mask-RCNN~\cite{he2017mask}.
The advantage is that segmentation mask does not have the problem of not aligning with the orientation of the person. They divided the images into a central region and a peripheral ring. The peripheral ring is further divided into three sectors, which are then warped into rectangles and stacked together to form one square image. The detection is performed using two separate detectors, one for the central region and the other for the outer region.
Fuertes \etal proposed a Grid of Spatial-Aware Classifiers~\cite{del2021} based on Deep Neural Networks (\textbf{GSAC-DNN})~\cite{fuertes2022people}.
A feature map is generated by a ResNet-32~\cite{he2016deep} backbone.
It is fed to a 2D grid of simple classifiers consisting of a convolution layer and a linear layer.
The location of the person is calculated based on the confidence scores of the classifiers.
GSAC-DNN is end-to-end trainable, however it can only detect the general position of the person but not a bounding box.
The work of \textbf{Callemein \etal}~\cite{callemein2019anyone} is intended for occupancy detection in meeting rooms or for flex-desking, yet the detection results are still presented as bounding boxes.
They use extremely low resolution images of \(96\times96\)\,pixels to preserve privacy.
To compensate for the information loss caused by the low resolution, they implemented a temporal interlacing kernel, which combines multiple consecutive frames into one high resolution feature map.
Their network is able to run on embedded systems such as the Raspberry Pi 3B~\cite{raspi3b} at \SI{0.77}{fps}.
Pais \etal used a deep Q-Net (DQN)-based~\cite{mnih2015human} network and the camera calibration parameters to perform person detection and predict the 3D position of the person in the world coordinate~\cite{pais2019omnidrl}.
The employment of reinforced learning is quite unusual.
Their implementation is named \textbf{OmniDRL}.

\subsection{Object detection}
\label{subsec:objectD}
Object detection in omnidirectional images has not been widely researched.
One reason is that it is not as useful as person detection. Another reason is the lack of data for training.
\textbf{Scheck \etal} created the THEODORE dataset~\cite{scheck2020learning} to solve this issue.
THEODORE contains five classes besides person: armchair, chair, table, TV and wheeled walker.
They trained SSD~\cite{Liu2016ssd}, R-FCN~\cite{dai2016rfcn} and Faster R-CNN~\cite{ren2015fasterrcnn} using this dataset and tested the trained networks on FES dataset.
The mAP for all six classes reached 0.613 with Faster R-CNN.
They used THEODORE to further train the anchorless CenterNet~\cite{duan2019centernet} for object detection~\cite{scheck2021uda}.
In this work, they introduced unsupervised domain adaptation (UDA) to bridge the gap between synthetic image domain and real-world image domain, which is typically used in semantic segmentation.
CenterNet was extended with two methods of UDA: entropy minimization (EM)~\cite{vu2019advent} and maximum squares loss (MSL)~\cite{chen2019domain}.
The unlabeled target dataset used in \cite{scheck2021uda} is CEPDOF.
With the UDA-extended CenterNet they raised the mAP on the same FES dataset to 0.690 and doubled the inference speed at the same time.
Another reason that object detection in omnidirectional view is underexplored is perhaps the very limited use cases.
However, this could still be useful for accomplishing complex tasks with omnidirectional images, such as action recognition for smart monitoring systems, for example, in \cite{seidel2018auxilia}, a system is built for monitoring elderlies with dementia.

\section{Human pose estimation and activity recognition}
\label{sec:pose}

Human pose estimation (HPE) refers to the process of finding the joints of a person and connecting them into a skeleton.
Pose estimation is the second most researched application of omnidirectional images.
In this section we take a look at the approaches for 2D and 3D pose estimation.
We also check out the researches for Human Activity Recognition (HAR), which often follows pose estimation.

\subsection{Pose estimation with overhead fisheye camera}
\label{subsec:pose}
\textbf{Georgakopoulos \etal}~\cite{delibasis2016geodesically,georgakopoulos2018pose} employ a 3D human model to create a dataset of binary silhouettes, which are rendered through the calibration of a fisheye camera.
The CNN is trained to differentiate between the pre-set postures, rather than estimate the joint positions.
\textbf{Denecke and Jauch}~\cite{denecke2021verification} use the 3D point cloud calculated by the smart sensor and prior knowledge of the human body to estimate the joint positions.
The results of this method are restricted by factors such as the mounting position of the camera and differences between each individual body.
The inference speed is limited by the speed of the smart sensor.
\textbf{Heindl \etal}~\cite{heindl2019large} generated rectilinear views of the area that contains human in an omnidirectional image, thus the person appears upright in the virtual view.
OpenPose~\cite{cao2017realtime} is then applied to this virtual image to perform pose estimation.
They used a pair of calibrated fisheye cameras to get two skeletons, which then are combined into a 3D skeleton by using Direct Linear Transform (DLT)~\cite{hartley1992stereo} in the rectilinear views.

Though not using omnidirectional images, the following two works perform HPE for the top-view.
Haque \etal train CNN and LSTM~\cite{hochreiter1997lstm} to achieve view-point invariant 3d pose estimation on a singular \emph{depth image}~\cite{haque2016recurrent,haque2016vpinvariant3dhpe}.
Garau \etal achieve viewpoint-invariant 3D HPE with a capsule auto-encoder named DECA~\cite{garau2021deca} on depth and RGB images, namely ITOP~\cite{haque2016vpinvariant3dhpe} and PanopTOP31K datasets.

\subsection{Egocentric 3D pose estimation}
\label{subsec:egocentric}
Egocentric pose estimation is a special case of using fisheye cameras in the top-view.
The camera is not mounted over the person, instead, it is mounted with an apparatus on the head of the person with a small horizontal distance.
Visible to the camera is the front or the frontal side of the body, as well as the peripheral of the person.
\textbf{EgoCap} is a dual fisheye camera set-up mounted on a bike helmet with either a T-shaped or a Y-shaped wooden frame~\cite{rhodin2016egocap}.
The cameras extrude about \SI{25}{\centi\meter} to the front of the carrier.
The authors created a dataset using a motion capture system to create the ground truth and projected the joint locations into the images from their set-up.
%They projected real-world backgrounds into the captured sequences and applied random gamma curve to augment the dataset.
With this dataset they finetuned ResNet101, which is pre-trained on MPII~\cite{andriluka20142d} and Leed Sports Extended Dataset~\cite{johnson2011learning}, to generate heatmaps for 18 joints.
The 3D skeleton is constructed in real-time from the 2D skeleton and a 3D body model, which must be adjusted for each user.
\textbf{Mo\textsuperscript{2}Cap\textsuperscript{2}}~\cite{xu2019mo} and \textbf{xR-EgoPose}~\cite{tome2019xr}\textbf{\,/\,SelfPose}~\cite{tome2020selfpose} are similar implementations with a single fisheye camera. % the former mounted it on a baseball cap, and the latter mounted it on a VR headset.
Both works developed their own synthetic datasets for training.
Mo\textsuperscript{2}Cap\textsuperscript{2} used one branch of CNN to generate heatmaps of joints for the whole body and another branch for the zoomed-in lower body.
With the help of a CNN that estimates the distance between the joints and the camera, the joints are finally projected into the 3D coordinates.
The calibration information of the cameras are essential for accurate 3D pose estimation.
xR-Egopose used ResNet101 for joint heatmap generation.
A lifting module takes the heatmaps as input and regresses the 3D pose from them as well as outputs the 2D heatmaps in high resolution.
\textbf{Wang \etal} proposed in \cite{wang2021estimating} a method for estimating not only the local pose, but also the global pose, which means the 3D joint positions in the world coordinate system are estimated.
Their pipeline makes use of image sequences instead of inferencing on a single frame.
At the same time, they utilized motion prior, which is learned from AMASS dataset~\cite{mahmood2019amass}, to reduce temporal jitter and unrealistic motions from the estimated poses.
Their set-up is similar to Mo\textsuperscript{2}Cap\textsuperscript{2} and xR-EgoPose by mounting a single fisheye camera onto a helmet.
In \cite{wang2022estimating} they further proposed to additionally use an external camera for weakly supervised training.
Their dataset for this task is named EgoPW.
\textbf{EgoGlass}~\cite{zhao2021egoglass} is more extreme in terms of minimizing the apparatus size.
They mounted two cameras to a normal eyeglass, each of which records a side of the body.
The pose estimation is solved in the stitched-together image.
We also notice \textbf{Cha \etal}~\cite{cha2018towards,cha2021mobile} used similar set-ups for their implementation, but not fisheye cameras.

\subsection{Action recognition}
\label{subsec:action}
\textbf{Li \etal}~\cite{li2020weakly} proposed to perform action recognition in top-view fisheye camera images.
They first use Mask-RCNN to find the spine lines of standing persons in the image.
The cross point of the spine lines are deemed the optical center of the fisheye camera and the spherical image is dewarped into a panoramic image around it.
Camera calibration information is not necessary in this process and the panoramic image is set to a pre-defined size.
The authors use Mask-RCNN to perform person detection in the panoramic image and max pool the bounding boxes across 16 frames in each clip to form the ROIs.
A 3D ResNet is used for action recognition through the 16 frames.
A binary mask, which is generated from the ROIs, is multiplied with the feature maps from the 3D ResNet to reduce calculation cost.
They used Multi-instance Multi-label Learning (MIML) to train a network for estimating scores for a series of actions.
This work is further developed by \textbf{Stephen \etal}~\cite{stephen2021hybrid} by adding a second parallel pipeline for persons in the central area.
Instead of using the panoramic view, this pipeline directly generates stacked feature maps for each person in the omnidirectional image, in which person detection is performed by RAPiD~\cite{duan2020rapid}.

\section{Other applications}
\label{sec:other}

Except for the intensively researched topics, multiple applications of omnidirectional images exist, mainly due to its wider FOV.
Researchers have applied deep learning to these applications, however, DL does not stand in the focus in these applications.
Much effort is given to solve the unique challenge of the equidistant projection of fisheye cameras, as well as other domain-specific problems.

Laurendin \etal proposed to use a top-view camera for \textbf{anomaly detection in train door area for autonomous trains}~\cite{laurendin2021hazardous}.
They created a dataset, in which the door area of a train is simulated, and systematically annotated it, see \cref{subsec:realD}. They adapted the network in \cite{nguyen2019anomaly}. The results are inconclusive.
Kim \etal proposed to use multiple top-view fisheye cameras for \textbf{parking lot surveillance} to determine vehicle positions~\cite{kim2020external}.
The cameras were first calibrated using RANSAC to obtain their intrinsic and extrinsic parameters.
SegNet\cite{badrinarayanan2017segnet} was used to generate segmentation masks for vehicle detections.
They developed a method to estimate the actual size of the vehicle based on the calibration parameters of the camera and the generated segmentation mask.
To get precise groundtruth data, the authors built a \(1/18\)-scale test bench using model cars and wooden frames.
Their method was tested with an average distance error of \SI{0.24}{\meter} (scaled to real-life) for vehicle position estimation and an average direction error of \SI{4.8}{\degree} for vehicle moving direction.
Akai \etal used a fisheye camera for \textbf{grape bunch counting}~\cite{akai2021distortion}. 
In contrary to other examples reviewed in this paper, this work uses the bottom-up view instead of the top-down view.
But this is the same approach in essence.
The difference of viewpoint is just because the ROI is over the head instead of on the ground.
Another important application is \textbf{indoor livestock monitoring}, such as body segmentation, identification, behavior recognition.
Using top-view omnidirectional cameras in such farming areas, which are usually large and densely packed with animals, provides an unoccluded view with minimum number of cameras.
Chen \etal~\cite{chen2021behaviour} provide a comprehensive review of this application area, including the use of top-view omnidirectional cameras and deep learning.
Li \etal performed \textbf{3D room reconstruction} using a single top-view fisheye camera~\cite{Li2019roomreconstruction}.
They used RefineNet~\cite{lin2017refinenet} for semantic segmentation of the room to aid structural line selection.
The final result of their method is a cuboid representation of the room.
% \subsection{Indoor Person Localization}
% \cite{yang2018ego}

% \subsection{Fall Detection}
% \label{subsec:fall}
% \cite{kottari2019real}\cite{siedel2020contactless}\cite{nguyen2021incorporation
% \input{sec/architecture}
\section{Conclusion}
\label{sec:concl}
We can conclude from our survey that the main application areas of top-view omnidirectional imaging are surveillance and AAL.
Researchers have created a considerable amount of data to facilitate the development of deep learning algorithms.
With this, the implementations of deep learning algorithms show very promising results.

An obviously underexplored research area is human pose estimation.
Though it has been greatly advanced for perspective images, the transfer to omnidirectional images is slow due to the high expense related to collecting human keypoints data with reliable ground truth~\cite{h36m_pami}.
The recent development in novel-view synthesis such as A-Nerf~\cite{su2021anerf} and HumanNerf~\cite{weng_humannerf_2022_cvpr} could be the solution to this problem.
More researches in this area will benefit further applications, such as fall detection, where only rule-based methods have been explored~\cite{delibasis2016falldetection,kottari2019real,siedel2020contactless,nguyen2021incorporation}.
Another possible research direction could be the usage of network architectures that are specifically developed for the geometry of fisheye images and related projections, such as spherical CNNs~\cite{cohen2018sphericalcnns} and SphereNet~\cite{coors2018spherenet}.

%%%%%%%%% REFERENCES
{\small
\bibliographystyle{ieee_fullname}
\bibliography{surveybib}
}

\end{document}